\documentclass[a4paper,twoside]{article}

\usepackage{epsfig}
\usepackage{subcaption}
\usepackage{calc}
\usepackage{amssymb}
\usepackage{amstext}
\usepackage{amsmath}
\usepackage{amsthm}
\usepackage{multicol}
\usepackage{pslatex}
\usepackage{apalike}
\usepackage{algorithm2e}
\usepackage[bottom]{footmisc}

\usepackage{graphicx} 
\usepackage{tabularray}
\usepackage{multirow}
\usepackage{mathtools}
\usepackage{caption}
\usepackage{subfiles} 
\usepackage{SCITEPRESS}     

\begin{document}

\title{Privacy-Preserving Face Recognition in Hybrid Frequency-Color Domain}

\author{\authorname{Dong Han\sup{1,2}\orcidAuthor{0000-0002-7782-3457}, Yong Li\sup{1,}$^{\ast}$\orcidAuthor{0000-0002-6920-0663} and Joachim Denzler\sup{2}\orcidAuthor{0000-0002-3193-3300}}
\affiliation{\sup{1}Huawei European Research Center, Riesstraße 25, 80992 München, Germany}
\affiliation{\sup{2}Computer Vision Group, Friedrich Schiller University Jena, Ernst-Abbe-Platz 2, 07743 Jena, Germany}
\email{\{dong.han2, yong.li1\}@huawei.com, \{dong.han, joachim.denzler\}@uni-jena.de}
}


\abstract{Face recognition technology has been deployed in various real-life applications. The most sophisticated deep learning-based face recognition systems rely on training millions of face images through complex deep neural networks to achieve high accuracy. It is quite common for clients to upload face images to the service provider in order to access the model inference. However, the face image is a type of sensitive biometric attribute tied to the identity information of each user. Directly exposing the raw face image to the service provider poses a threat to the user's privacy. Current privacy-preserving approaches to face recognition focus on either concealing visual information on model input or protecting model output face embedding. The noticeable drop in recognition accuracy is a pitfall for most methods. This paper proposes a hybrid frequency-color fusion approach to reduce the input dimensionality of face recognition in the frequency domain. Moreover, sparse color information is also introduced to alleviate significant accuracy degradation after adding differential privacy noise. Besides, an identity-specific embedding mapping scheme is applied to protect original face embedding by enlarging the distance among identities. Lastly, secure multiparty computation is implemented for safely computing the embedding distance during model inference. The proposed method performs well on multiple widely used verification datasets. Moreover, it has around 2.6\% to 4.2\% higher accuracy than the state-of-the-art in the 1:N verification scenario.}

\onecolumn \maketitle \normalsize \setcounter{footnote}{0} \vfill

\renewcommand{\thefootnote}{\fnsymbol{footnote}}
\footnotetext[1]{Corresponding author}

\section{\uppercase{Introduction}}

With the development of computational power and advanced algorithms, state-of-the-art (SOTA) face recognition (FR) models have achieved quite high accuracy in public open-source datasets. However, privacy concerns raise attention with the advance of artificial intelligence (AI). Since the deep learning-based method needs enormous amounts of facial data, it has more risks in terms of sensitive information leakage. Therefore, it is necessary to develop a mechanism to protect privacy information while maintaining the high utility of the FR system.

The acquisition of large-scale face images from the public through various service providers or organizations is becoming an important concern. The storage of face images is restricted, especially for the original ones, for the consideration of the potential misuse of analyzing personal sensitive information such as ethnicity, religious beliefs, health status, social status, etc. Hence, since most face recognition systems require access to the raw face image, those concerns restrict traditional FR usage to a certain extent. 

Another risk of FR is in face embeddings since they can be considered as one type of biometric data. The embedding contains the information that is extracted from the face, which can be used for identification. In the scenario of multiple stored face embedding datasets computed by the FR system, if there are the same identities existing in different datasets, the FR model can be used for identifying the common identities (1:N) or simple cross-authentication. The demographic information (e.g., sex, age and race) from the target face embedding dataset can be inferred by re-identification attacks with the help of a corresponding public accessible database \cite{fabian2020anonymizing}. Most convolutional neural networks (CNNs) FR models rely on output embeddings to distinguish faces from different identities. By simplifying the solution of the embedding vector to a normalized face space, an end-to-end decoder is trained by the texture and landmark of the face image to recover the predictive normalized face (which is a front-facing and neutral-expression face) \cite{cole2017synthesizing}. Face embeddings of the same face computed from different models should be independent and non-correlated intuitively. However, researchers \cite{mcneely2022canonical} claim it is possible to find a linear transformation to map face embedding between two models. It poses a threat that one can use the mapping function between two networks to infer the interest identity in another embedding database.

In this paper, we focus on reducing the privacy information in two stages of the FR system which involve the input of the feature extractor and the output face embedding. Most frequency-based FR methods have high-dimensional inputs without utilizing any color information \cite{ji2022privacy,wang2022privacy,mi2023privacy}. We propose a frequency fusion method that selects the most important coefficients across the channel to discard redundant information. The sparse visual color information with fused frequency information from the proposed hybrid frequency-color domain is the raw input to the feature extractor before adding perturbation. Then, the output face embeddings are projected into an identity-protected space. The separation of embeddings from different identities is enlarged during the mapping process. Unlike most previous works of the FR which focused on 1:1 verification as the main performance evaluation, accuracy is also reported in the 1:N verification scenario in our work.


As a summary, the main contributions of this research work are the following:
\begin{itemize}
  \item We propose a frequency fusion approach which enables dimensionality reduction in the existing frequency domain-based face recognition model.
  \item A hybrid frequency-color information fusion method is designed to improve recognition accuracy by combining sparse color and frequency information together without revealing much visual information.
  \item The identity-specific separation characteristic of the face embedding protection method is extensively investigated in 1:N face verification.
  \item Secure multiparty computation (SMPC) is applied to embedding distance calculation to further enhance the robustness against the reverse attack of face embeddings.
\end{itemize}


\section{\uppercase{Related Work}}



\subsection{Privacy-Preserving Face Recognition}
The privacy-preserving face recognition (PPFR) technique attracts more and more attention from both academia and industry. In general, users with limited computation power access FR service by uploading their face image to a service provider having a pre-trained model based on similar datasets. In this scenario, the user's face image is directly exposed to the service provider during the inference stage. Therefore, when the privacy of facial information is concerned or the user is simply not willing to share the original face image, the utility of such face recognition will be affected. For the privacy protection of face recognition, the traditional way is to add certain distortions such as blur, noise and mask to face images for reducing the privacy \cite{korshunov2013using}. These naive distortions produce unsatisfying recognition performance and the original face is relatively easy to be reconstructed. Homomorphic encryption is an encryption technique that ensures data privacy by encrypting the original data and enabling computations to be performed on the encrypted data \cite{ma2017secure}. Nevertheless, the encryption method necessitates significant additional computational resources. Differential privacy (DP) is another typical way to protect privacy by adding perturbation to the original or preprocessed face images \cite{chamikara2020privacy}. In order to strengthen information privacy for face recognition, the researchers \cite{wang2022privacy} propose the privacy-preserving FR in frequency-domain (PPFR-FD), which selects pre-fixed subsets of frequency channels and implements operations such as channel shuffling, mixing, and discording the lowest frequency channel. The learnable DP noise is introduced to reduce the visual information in the frequency domain while maintaining high utility of recognition \cite{ji2022privacy}.

\begin{figure*}[t]
\centering
\includegraphics[scale=0.12]{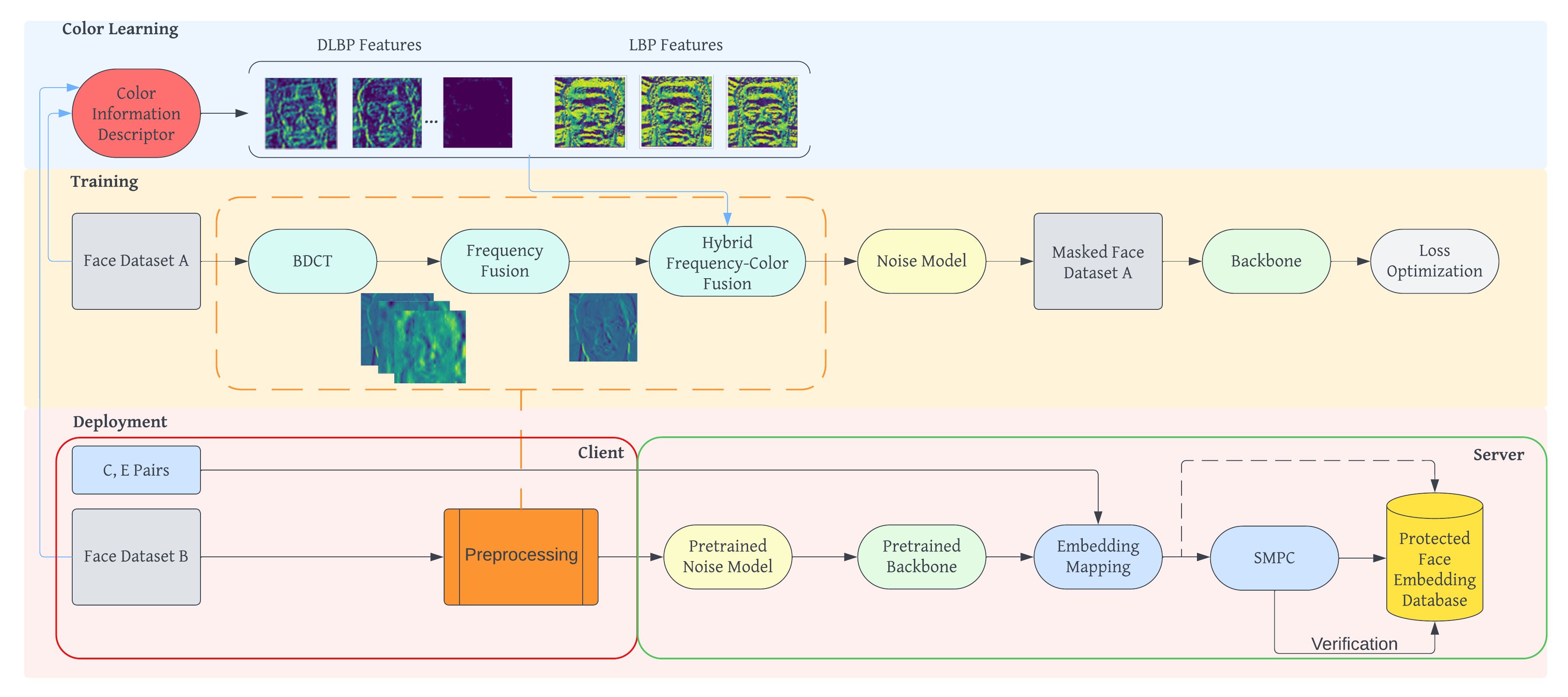}
\caption{Overview of proposed privacy-preserving FR framework in hybrid frequency-color domain.}
\label{fig:workflow}
\end{figure*}

\subsection{Frequency Domain Learning}
Discrete cosine transform (DCT) is a powerful transformation technique in image processing, commonly used in JPEG encoding \cite{wallace1991jpeg}. DCT represents images in the form of cosine waves. For human observers, the major visual information inside the image is contributed by the low frequency, while the high frequency only contains subtle visual information. Image data is the major input format for most CNNs. By accelerating neural network training, the traditional RGB image inputs can be replaced by the corresponding DCT coefficients to speed up the process \cite{gueguen2018faster}. The frequency representation has been used for different image tasks including image classification \cite{Ulicny2017OnUC} and segmentation \cite{lo2019exploring}. Avoiding using all frequency channels in DCT representation and selecting a small number of low frequency channels for training is also possible to maintain the relative high accuracy \cite{xu2020learning}.


\subsection{Color Domain Learning}
Color information is considered to be the primary and significant element within an image since it is strongly associated with objects or scenes. Consequently, it is widely recognized and utilized as a fundamental feature in the fields of image recognition and retrieval. The local binary patterns (LBP) technique is designed for texture description \cite{ojala1996comparative} and it has proven to be highly discriminating for FR due to different levels of locality. The face image is partitioned into several patches, and texture descriptors are retrieved from each region separately. Then the descriptors are combined to provide a comprehensive depiction of the facial features. The corresponding histograms from descriptors can be used as feature vectors for the FR model \cite{ahonen2006face}. However, LBP only extracts texture information from the image but color information is ignored. The goal of color-related local binary pattern (cLBP) \cite{xiao2020color} is to learn the most important color-related patterns from decoded LBP (DLBP) so that color images can be recognized.
In their work, the LBPs are computed from the proposed color space relative similarity space (RSS) besides from RGB channels. Then the LBP is converted into the DLBP by the decoder map.


\subsection{Face Embedding Protection}
There are two main types of methods for face embedding protection: handcrafted-based and learning-based algorithms. Handcrafted-based methodologies employ algorithmically defined transformations to convert face embeddings into more secure representations \cite{pandey2015secure,drozdowski2018benchmarking}. Normally, a learning-based method is associated with the feature extractor in the network. CNN-based protection method learns a mapping function to convert the extracted feature vector to maximum entropy binary (MEB) codes \cite{kumar2016deep}. Bioconvolving method \cite{abdellatef2019cancelable} is able to generate cancelable biometric embedding directly based on the deep features from CNN.

PolyProtect \cite{hahn2022towards} is one type of handcrafted approach that is able to convert original face embeddings to protected ones by using multivariate polynomials. It can be directly used as an independent module after the feature extractor. It is incorporated into our proposed framework.

\section{\uppercase{Methodology}}

In this section, the proposed FR framework is illustrated in Figure \ref{fig:workflow}.

The whole framework consists of three stages including color learning, training and deployment. In the color learning stage, the DLBP and LBP are extracted to represent the local color details of the original image, which are later mixed with fused frequency information. In the training stage, the size of the DCT image is $[H,W,C]$ while $C$ is the number of frequency channels. Then, the frequency fusion module reduces the channel number by $C/3$ before adding the pixel-wise DP noise through the noise model. In the deployment stage, the client uploads a perturbated frequency-color hybrid representation to the server side for feature extraction. Besides, compared to the traditional FR system, the embedding mapping based on PolyProtect is incorporated to protect original face embeddings in order to enlarge the distance among different identities by the identity-specific pairs C, E defined by the client. Lastly, the SMPC serves as extra protection to safely compute the distance during the verification stage.


\subsection{Block Discrete Cosine Transform (BDCT)}

\begin{figure}[hbt!]
\centering
\includegraphics[scale=0.22]{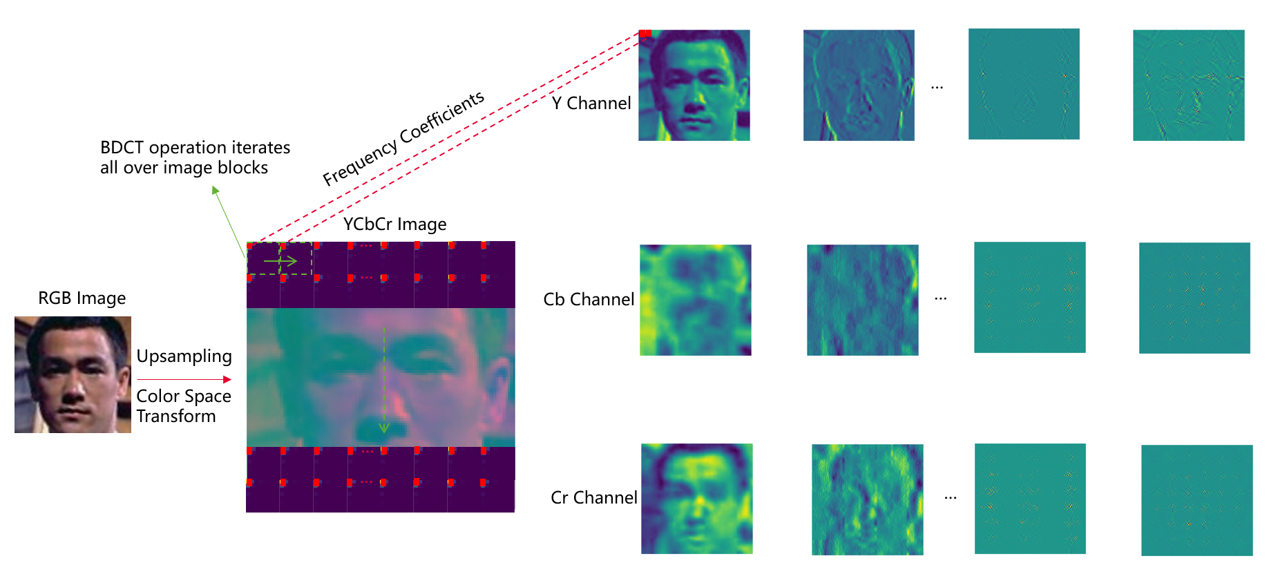}
\caption{BDCT operation on a face image.}
\label{fig:BDCT}
\end{figure}

BDCT splits images into different blocks and converts them into frequency representations, and it can be used for both color and grayscale images. As depicted in Figure \ref{fig:BDCT}, the original RGB image is upsampled by 8 times and transformed into YCbCr color space before the BDCT operation in order to keep the global face structure in each frequency channel. In this case, the block size of DCT is in $8\times8$ pixels, and therefore each channel derives 64 DCT images. The frequency coefficients range from -1024 to 1023. The Y component contains the most obvious (in terms of human perception) grayscale information about the content in the image while Cb and Cr carry information about the colors. Besides, the output DCT images in the first column are the direct currents (DCs) which represent the lowest frequency information.

\begin{figure}[hbt!]
     \centering
     \begin{subfigure}[b]{0.22\textwidth}
         \centering
         \includegraphics[width=\textwidth]{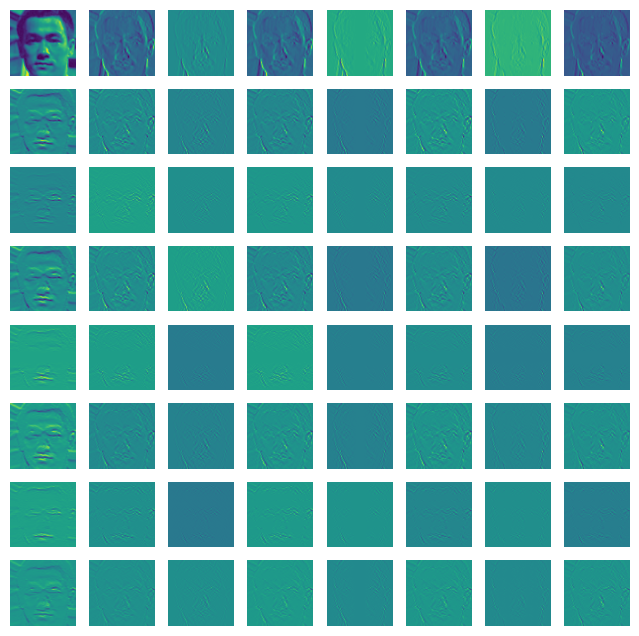}
         \caption{}
         \label{fig:DCT_image_64_Y}
     \end{subfigure}
     \hfill
     \begin{subfigure}[b]{0.22\textwidth}
         \centering
         \includegraphics[width=\textwidth]{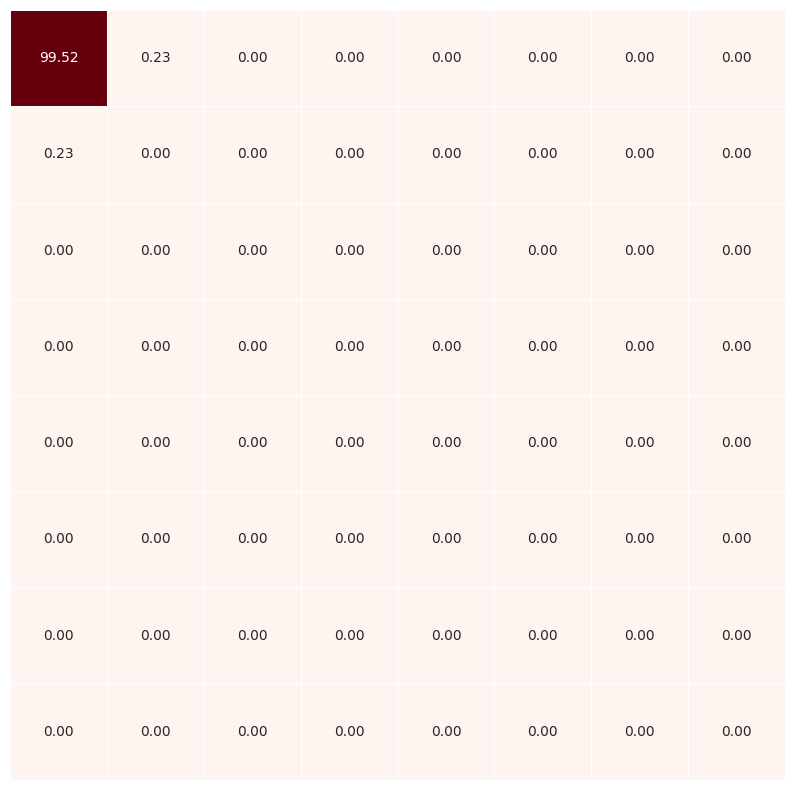}
         \caption{}
         \label{fig:DCT_energy_64_Y}
     \end{subfigure}
         \caption{DCT face images in Y channel and the DCT energy.}
         \label{fig:DCT_image_and_energy_64_Y}
\end{figure}

The DCT coefficients in the top-left channel correlate to the lowest frequency channel shown in Figure \ref{fig:DCT_image_64_Y}. Away from the channel in all directions (horizontal, vertical and diagonal), the coefficients correlate to higher frequencies, the right-bottom channel corresponds to the highest frequency. The low-frequency channels contain more visual structure than the high-frequency ones. The frequency image energy G is defined as:

\begin{equation}\label{eq:DCT_energy}
G = \sum_{i}^{}\sum_{j}^{}I(i,j)^2 
\end{equation}

where I is the DCT representation and (i, j) denotes the position of each coefficient.

According to Equation \ref{eq:DCT_energy}, the energy of each frequency is calculated and the percentage of energy distribution among all channels is shown in Figure \ref{fig:DCT_energy_64_Y}. The lowest frequency channel accounts for around $99\%$ of energy among all 64 channels.

\subsection{Frequency Fusion (FF)}
The dimensionality of the BDCT output is 189, even after dropping the DCs. It requires a huge amount of storage space if BDCT images are needed to be stored, and it also introduces more difficulty in the training stage for model convergence. In order to solve such a problem, we propose a frequency fusion scheme to reduce the dimensionality.

\begin{figure}[hbt!]
\centering
\includegraphics[scale=0.26]{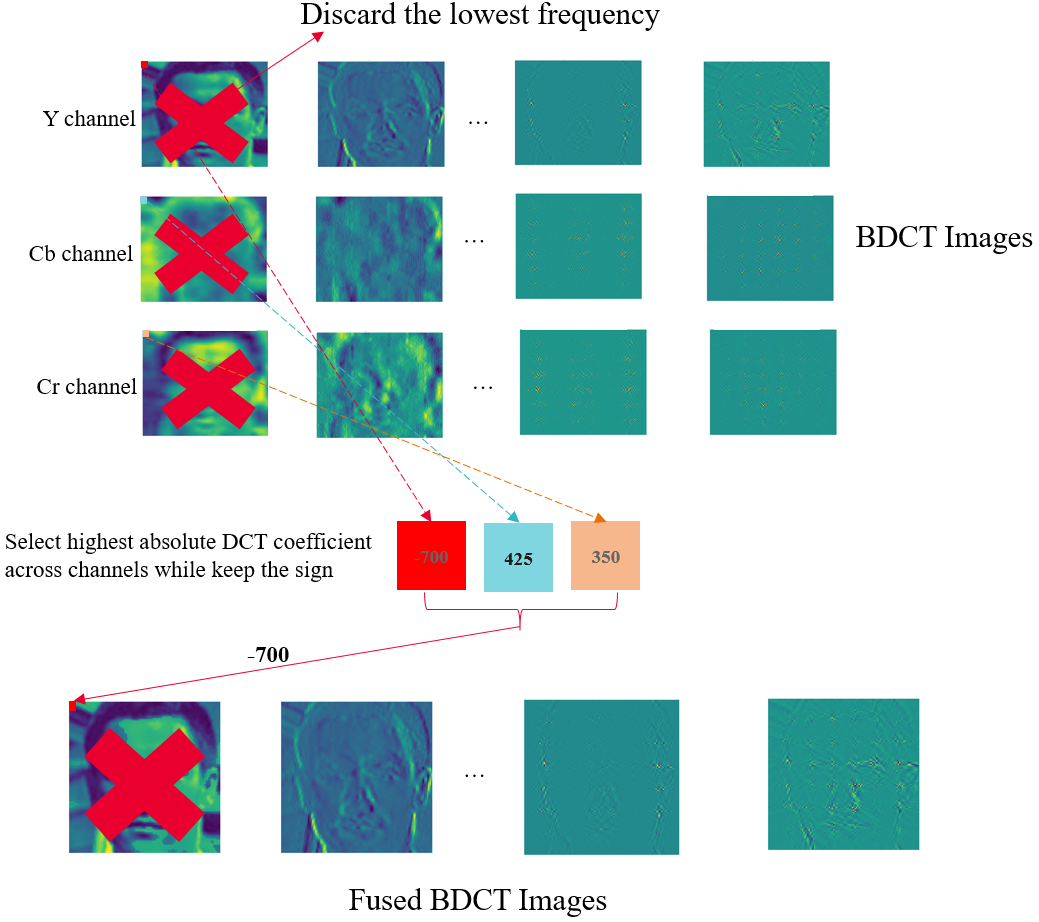}
\caption{Cross-channel frequency fusion on BDCT images.}
\label{fig:frequency fusion}
\end{figure}

Inspired by the previous study, the high absolute DCT coefficient indicates high importance for the visual structure. Therefore, it is possible to combine BDCT images across the channel. As illustrated in Figure \ref{fig:frequency fusion}, for the BDCT images in Y, Cb, Cr channels at the same level of frequency, the highest absolute value from the pixel in the same position is selected as the final coefficient. It is noticed that the corresponding sign of the selected value is kept. After the frequency fusion, the fused BDCT image contains only 63 channels which is three times less than the input dimension used in the paper \cite{ji2022privacy}.

\subsection{Color Information Descriptor}
As the only frequency information is used for recognition in the frequency domain, the color information from the original RGB image is completely dismissed, which hinders the recognition accuracy. The goal of the proposed color information descriptor is to extract and transfer color information into the representation without preserving much visual structure. Our implementation directly uses the decoded local binary pattern (DLBP) as a sparsity representation of color information, drawing inspiration from the work \cite{xiao2020color}. Additionally, the classical LBPs are also computed for comparison.

\begin{figure}[hbt!]
\centering
\includegraphics[scale=0.5]{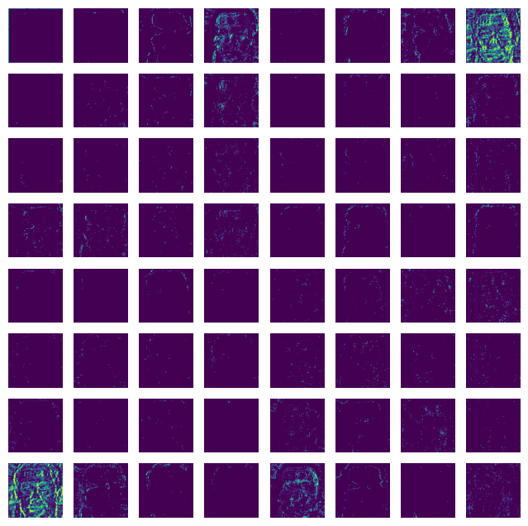}
\caption{DLBP features. They are computed based on the same example RGB image used in Figure \ref{fig:BDCT}.}
\label{fig:DLBP}
\end{figure}

As shown in Figure \ref{fig:DLBP}, only a few images contain certain ambiguous visual information. Most facial details and contours are barely perceived in contrast with traditional LBP features.

\subsection{Hybrid Frequency-Color Fusion (HFCF)}

\begin{figure}[hbt!]
\centering
\includegraphics[scale=0.34]{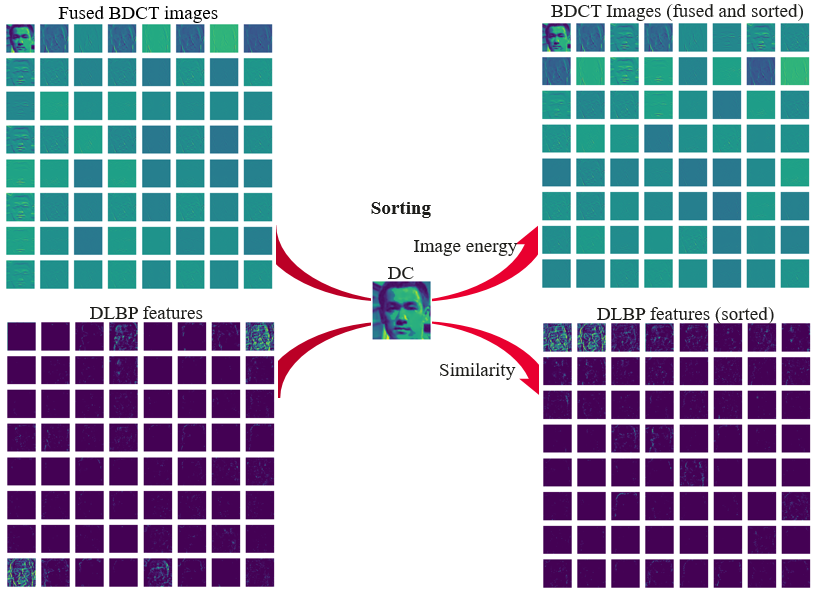}
\caption{Frequency-color sorting.}
\label{fig:frequency-color-soring}
\end{figure}

To the best of our knowledge, there is no explicit way to combine or mix the frequency and color information. For the frequency information, even though it is known that the frequency decreases from the upper left corner to the lower right corner, the exact order cannot be easily observed. Our method is quite intuitive, as shown in Figure \ref{fig:frequency-color-soring}. Firstly, the fused BDCT images are sorted according to the DCT image energy calculated by Equation \ref{eq:DCT_energy}. We sort output BDCT images in descending order and arrange them row by row. Secondly, the DLBP features are sorted by checking the similarity (e.g., Euclidean distance) with respect to the DC component in fused BDCT images. The sorted DLBP features are also in descending order.

\begin{figure}[hbt!]
\centering
\includegraphics[scale=0.3]{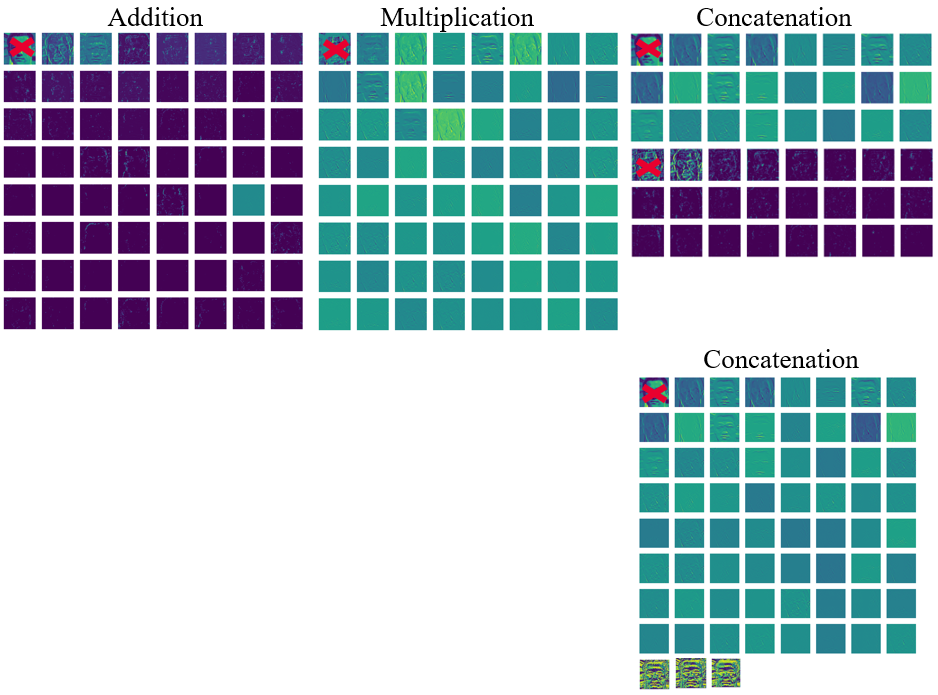}
\caption{Hybrid frequency-color information. The DC as well as the first DLBP feature are excluded. The four different fusions provide options for the model backbone input.}
\label{fig:hybrid-frequency-color-information}
\end{figure}

For sorted frequency and color information, we present the multiple naive fusion schemes (e.g., addition, multiplication, and concatenation). It is noticed that the LBP features are only used in concatenation since the LBP contains a relatively high visual structure which means it is less privacy-preserving compared with the DLBP features. The hybrid frequency-color information representations shown in Figure \ref{fig:hybrid-frequency-color-information} are the different input options for the model backbone before adding the DP noise.

In order to analyze the privacy-preserving quality, the visual similarity between the proposed hybrid frequency-color information and the original image is compared. Besides quantifying image compression quality, the peak signal-to-noise ratio (PSNR) is a useful metric of image similarity as well as the structural similarity index measure (SSIM). Moreover, SSIM can reflect a certain amount of human vision perceptual quality.

\begin{figure}[hbt!]
\centering
\includegraphics[scale=0.35]{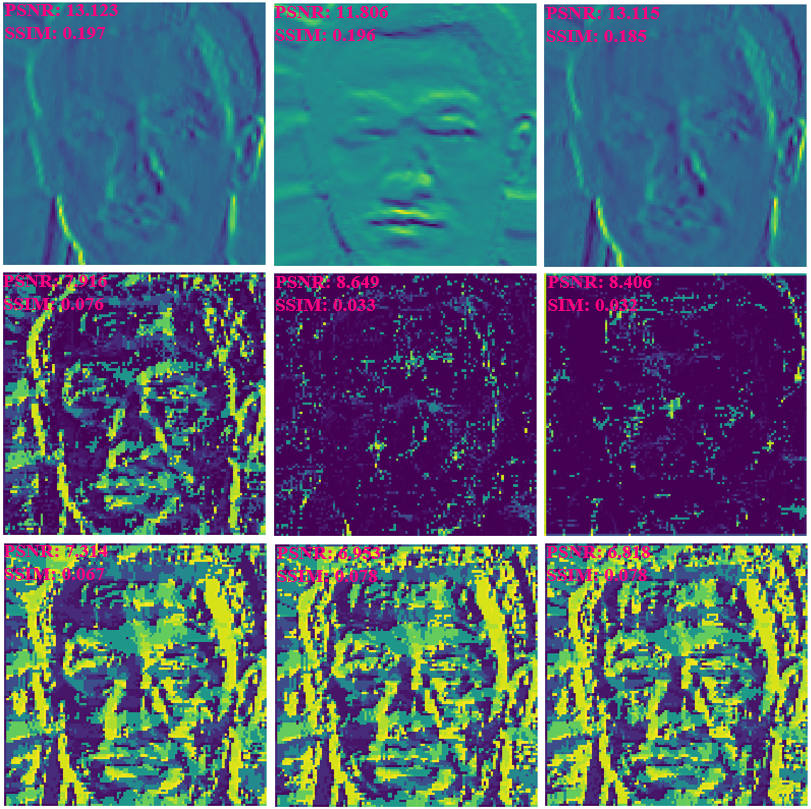}
\caption{PSNRs (db) and SSIM between the original RGB image (luma channel in YCbCr space is used for calculation) and the ones in frequency and color domains. The lower the value, the better the privacy-preserving. The first three sorted BDCT and DLBP images are shown in the first and second row; the LBPs on R, G, and B channel are shown in last row.}
\label{fig:PSNRs}
\end{figure}

As shown in Figure \ref{fig:PSNRs}, the DLBP and LBP have a lower PSNR than the ones in the frequency domain, which means they are more privacy-preserving. However, PSNR performs less well in evaluating the quality of images perceived by humans. It is obvious that DLBP images, especially the last two, contain less visual information compared with the ones from LBPs. According to the SSIM comparison, DCT representations contain more structural information than DLBP and LBP. Additionally, DLBP captures less information than LBP. It shows that DLBP is more difficult to be interpreted in terms of visual perception.

\subsection{Embedding Mapping}

Assuming the original face embedding $V$ = [$v_1$, $v_2$, ..., $v_n$], the protected face embedding is denoted as P = [$p_1$, $p_2$, ..., $p_n$]. As stated in PolyProtect, the mapping operation is achieved by following formula. For the first value in $p_1$:, 

\begin{equation} \label{PolyProtect1}
p1 = C_1*v_1^{e_1} + C_2*v_2^{e_2} + ... +C_m*v_m^{e_m}
\end{equation}

where C = [$c_1$, $c_2$, ..., $c_m$] and E = [$e_1$, $e_2$, ..., $e_m$] are vectors that contain non-zero integer coefficients. The first m values in V are mapped into $p_1$, then $p_2$ is calculated based on another m consecutive values after m. There is no obvious evidence to choose the range of C if the cosine distance metric is selected since it is not sensitive to magnitude changes. For the range of E, it is reasonable to avoid large numbers since face embedding consists of small floating point values, while large powers wipe out certain embedding elements. We keep m = 5 and E in the range [1, 5] as the author suggested. Since we aim to generate a unique C vector for each identity, a large C range [-100, 100] is used in our setting.

Another important parameter is named overlap, which indicates the number of common values from V that are used in the computation of each value in P. For instance, $v_6$$\sim$$v_{10}$ are values for the calculation of $p_2$ when overlap = 0, while $v_5$$\sim$$v_9$ is selected in case of overlap = 1, by repeating use the $v_5$ that already being used in computation of $p_1$. For reversibility, when overlap is 4, it has a high possibility to reverse target P to an approximated V' when the formula and all parameters are known \cite{hahn2022towards}.

We focus on handling this issue in our SMPC-based method in Section \ref{sec:SMCP}. Besides, all the experiments in the original PolyProtec paper are mainly performed for 1:1 verification. In our work, the PolyProtect is tested in 1:N verification in Section \ref{sec:experimental results}.


\subsection{Secure Multiparty Computation (SMPC)}\label{sec:SMCP}

\begin{figure}[hbt!]
\centering
\includegraphics[scale=0.19]{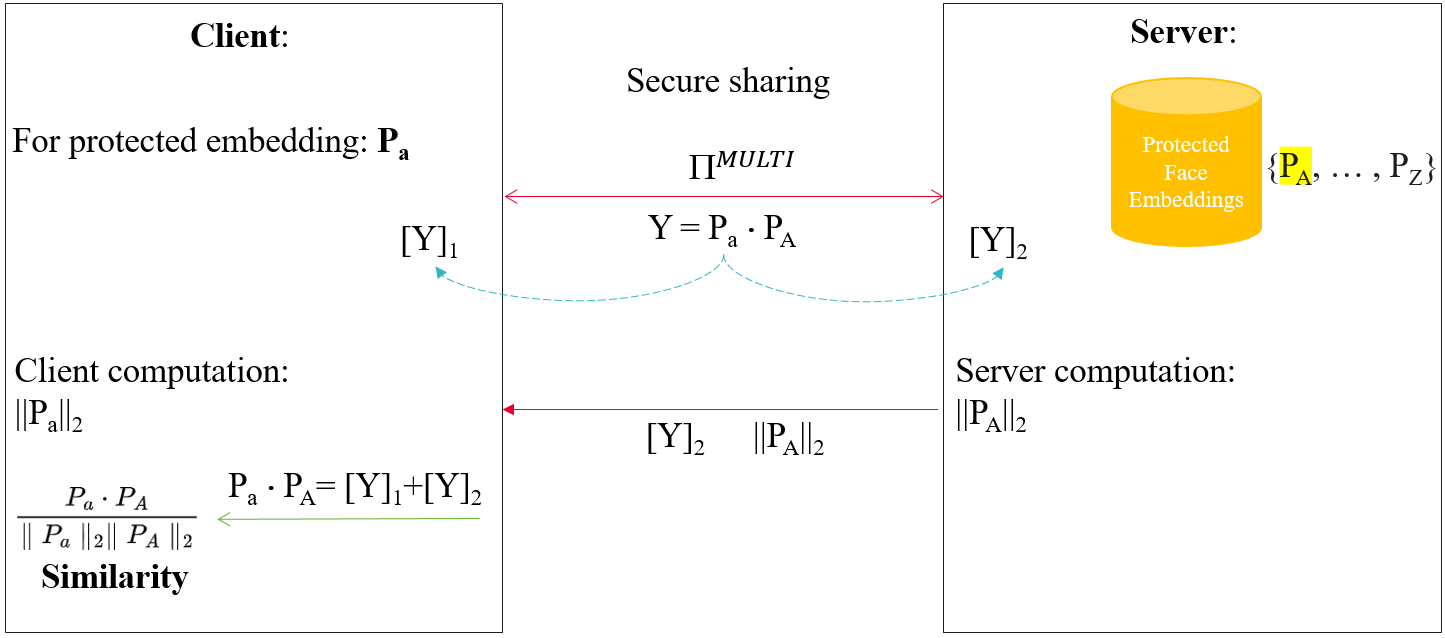}
\caption{SMPC for embedding distance computation.}
\label{fig:SMPC}
\end{figure}

Through the integration of SMPC with other cryptography methodologies \cite{evans2018pragmatic}, it becomes possible to secretly verify the encrypted biometric attributes of a user with the previously provided data. In order to tackle the issue of using overlap = 4 in PolyProtect, we propose SMPC-based similarity computation on protected embeddings, as shown in Figure \ref{fig:SMPC}.

During the verification stage, the secure sharing protocol $\Pi^{MULTI}$ is established when comparing the protected embedding $P_a$ and the enrolled embedding $P_A$ from the database. The dot product results Y($P_a$$\cdot$$P_A$) are split into $[Y]_1$ and $[Y]_2$ for the client and server, respectively. Then $[Y]_2$ and the L2 norm of the enrolled embedding $P_A$ are sent to the client. Lastly, the dot product can be calculated by $[Y]_1$ + $[Y]_2$. 
With all elements, the client is able to compute the cosine similarity between the protected embedding and each enrolled embedding in the database. The corresponding cosine distance can also be easily derived from the cosine similarity. By avoiding sharing the raw protected embedding with the server, the risk of overlap 4 in PolyProtect is solved. Our implementation is based on CrypTen \cite{knott2021crypten} in PyTorch. For detailed information about secure sharing, we recommend referring to the work \cite{damgaard2012multiparty}.

\section{\uppercase{Experiments}}
\subsection{Dataset}
The VGGFace2 \cite{cao2018vggface2} dataset is utilized as a training dataset; it comprises 3.31 million photos of 9131 identities, with over 300 images for each identity and a wide range of posture, age, lighting and ethnicity. For the 1:1 verification, we explore efficient face verification benchmark datasets including Labeled Faces in the Wild (LFW) \cite{huang2008labeled}, Celebrities in Frontal Profile (CFP) \cite{sengupta2016frontal} In-the-Wild Age Database (AgeDB) \cite{moschoglou2017agedb} to check the model performance. Besides the most widely used, we also report the performance of different models on datasets (e.g., Cross-Age LFW (CALFW) \cite{zheng2017cross} and Cross-Pose LFW (CPLFW) \cite{zheng2018cross}) with large pose and age variations to test the model generalization and robustness. For the 1:N verification, the customized dataset that was constructed from MS-Celeb-1M \cite{guo2016ms} is used. It consists of two parts: the gallery dataset and the query dataset. The
The former only contains 1 image per identity, as the selected images are used for generating the enrolled embedding database. The latter has five images for each identity, while the one used in the gallery dataset is excluded during the selection process.


\subsection{Implementation Details}
For the input RGB face image, it is aligned (based on MTCNN \cite{zhang2016joint}) and resized into 112 $\times$ 112 pixels. Then the image is upsampled by 8 times (through bilinear interpolation) in order to keep global visual structure in the frequency domain. The upsampled image is transferred into YCbCr color space before performing BDCT with a block size of 8 × 8 pixels. The partial implementation is based on TorchJPEG \cite{ehrlich2020quantization}.

The initial BDCT image has 192 channels in total. After dropping the DCs and applying the proposed frequency fusion scheme illustrated in Figure \ref{fig:frequency fusion}, the fused BDCT image only contains 63 channels. Then the LBP and DLBP features are computed based on the original RGB face using the proposed color information descriptor. The LBP is computed by converting each pixel into a binary number (8 digits) in comparison with its 8 neighbors. We calculate the LBP feature separately on each channel. The DLBP is computed by our implementation of LBP decoding \cite{xiao2020color}. Then, the proposed frequency-color sorting and hybrid frequency-color information scheme combine the fused BDCT image with LBP and DLBP features through addition, multiplication and concatenation, the detailed operation is shown in Figure \ref{fig:frequency-color-soring} and Figure \ref{fig:hybrid-frequency-color-information}. Before feeding the hybrid frequency-color information to the backbone, the learnable DP noise (the implementation according to paper \cite{ji2022privacy} is added. The baseline model is based on the ResNet-34 \cite{He_2016_CVPR} backbone. The same random seed is set in all experiments. The model is trained for 24 epochs with a batch size of 128. The stochastic gradient descent (SGD) optimizer is selected with 0.9 momentum and 0.0005 weight decay, respectively. For the loss function ArcFace \cite{deng2019arcface}, s is set to 64 and m is set to 0.3. All experiments are conducted on 2 NVIDIA Tesla 100 GPUs with the PyTorch framework.

\begin{table*}[t]
\caption{Comparison of 1:1 verification accuracy of different methods. DCTDP-FF denotes the fused frequency domain, which is applied to the baseline DCTDP. Concat, add and multi denote the different options in HFCF.}\label{tab:1:1 verification} \centering
\begin{tabular}{|c|c|c|c|c|c|c|c|}
  \hline
  Method (\%) & \# Channels & LFW & CFP-FP  & AgeDB & CALFW & CPLFW\\
  \hline
  ArcFace \cite{deng2019arcface} & 3 & 99.70 & 98.14 & 95.62 & 94.28 & 93.10\\
  DCTDP \cite{ji2022privacy} & 189 & 99.64 & 97.69 & 95.10 & 93.87 & 91.77\\
  DCTDP-FF & 63 & \textbf{99.60} & 97.69 & 94.95 & 93.25 & 91.83\\
  HFCF-LBP (concat) & 66 & 99.58 & 97.76 & 94.63 & \textbf{93.60} & \textbf{91.87}\\
  HFCF-DLBP (add) & 63 & 99.37 & \textbf{97.77} & 94.52 & 92.93 & 90.55\\
  HFCF-DLBP (concat) & 30 & 96.03 & 88.57 & 83.70 & 83.92 & 80.87\\
  HFCF-DLBP (concat) & 126 & 99.57 & 97.69 & \textbf{95.03} & 92.95 & 91.70\\
  HFCF-DLBP (multi) & 63 & 99.25 & 97.50 & 94.43 & 93.16 & 90.87\\
  \hline
\end{tabular}
\end{table*}

\subsection{Experimental Results}\label{sec:experimental results}

\subsubsection{1:1 Verification}

We compare the results with the SOTA baseline models: ArcFace is trained with unprotected RGB images and DCTDP is trained in a frequency domain protected by perturbation of DP noise. We test our proposed frequency fusion and hybrid frequency-color method on popular 1:1 verification datasets. The recognition accuracy is shown in Table \ref{tab:1:1 verification}. It is good to notice that our DCTDP-FF method can maintain high accuracy by only 63 frequency channels compared with the DCTDP which keeps 189. Besides, the color information from LBP and DLBP is helpful for improving accuracy in some cases. However, there are no obvious changes since LBP and DLBP are quite sparse. When only the first 15 channels are selected from frequency and DLBP in HFCF-DLBP (concat), it reduces the accuracy significantly. According to their relative high performance, HFCF-LBP (concat) and HFCF-DLBP (concat) with full channels are selected as the main methods to investigate in further experiments.

\subsubsection{1:N Verification}
The majority of the experiments in previous works were mainly conducted in a 1:1 verification setting to evaluate FR performance. However, 1:N verification is the most common situation in real-world application. Therefore, we also test our methods in a 1:N verification scenario. The enrolled embedding database is computed based on the images from the gallery dataset. Each image represents a unique identity and there are 85742 in total. In the inference stage, an image from a random identity is picked to compare the distance with the embeddings in the enrolled embedding database.

For a more fair 1:N verification accuracy measurement on different models, the randomness of the selection of query images from the query dataset is fixed. The mean accuracy is calculated for every 1000 query images. As shown in Table \ref{tab:1:N verification}, in original embedding case, the model only applied frequency fusion DCTDP-FF has lower accuracy than DCTDP in all three top rank predictions. The model with LBP features concatenated HFCF-LBP (concat) has better performance than the pure frequency fusion model. It also has higher accuracy than DCTDP. It is good to notice that the method with DLBP features concatenated has the best performance among all the models, and it achieves 2.6\%, 4.3\% and 4.2\% more accuracy than the SOTA baseline DCTDP. Therefore, our proposed hybrid frequency-color domain, especially the one based on DLBP features HFCF-DLBP (concat) has more useful information for recognition even though it only provides trivial visual representation of original face image. Such a characteristic is suitable for privacy preservation since most of the facial structure is concealed from the inputs of the backbone. As showing in Figure \ref{fig:PSNRs}, DLBP image has low SSIM value and only ambiguous contour is presented. 

\begin{table*}[t]
\caption{Comparison of 1:N verification accuracy on original embedding (the direct output of the backbone) and protected embedding (mapping by PolyProtect with overlap 4) of different methods. The retrieval rate are calculated for different top rank predictions.}\label{tab:1:N verification} \centering
\begin{tabular}{|c|c|c|c|c|c|c|c|}\hline
      Method (\%) & \# Channels & \multicolumn{6}{|c|}{Retrieval Rate}\\\cline{3-8}
       & & \multicolumn{3}{|c|}{Original Embedding} & \multicolumn{3}{|c|}{Protected Embedding}\\\cline{3-8}
       & & Rank=1 & Rank=5 & Rank=10 & Rank=1 & Rank=5 & Rank=10\\\cline{1-8}
      ArcFace \cite{deng2019arcface} & 3 & 87.8 & 93.8 & 95.3 & 95.2 & 97.3 & 98.1\\\hline
      DCTDP \cite{ji2022privacy} & 189 & 79.3 & 86 & 88.3 & 88.7 & 93.9 & 95.3\\\hline
      DCTDP-FF & 63 & 75 & 84 & 84.9 & 88.3 & 93.3 & 95.1\\\hline
      HFCF-LBP (concat) & 66 & 80.9 & 88.4 & 89.9 & 89.0 & 94.8 & 96.4\\\hline
      HFCF-DLBP (concat) & 126 & \textbf{81.9} & \textbf{90.3} & \textbf{92.5} & 91.2 & 95.7 & 96.8\\\hline
\end{tabular}
\end{table*}

Apart from the accuracy improvement brought by the proposed hybrid frequency-color information, the 1:N performance is further enhanced by the identity-specific embedding mapping. Based on empirical experiments on different C, E parameter settings, large range of E can degrade recognition accuracy while the range of C is suggested to be larger enough to generate different combinations (at least more than the number of identity). In terms of privacy preserving, since proposed FR system requires user-specific parameters C and E, it is difficult to access the FR system even the identity image is leaked. In the rank 1 prediction scenario, the accuracy is increased by 7.4\% for ArcFace, 9.4\% for DCTDP, 13.3\% for DCTDP-FF, 8.1\% for HFCF-LBP (concat) and 9.3\% for HFCF-DLBP (concat). Furthermore, the accuracy in the other two rank cases is raised to a considerable extent when compared to the accuracy computed using the original embedding setting.

\subsubsection{Hard-Case Query Performance}

\begin{figure*}[t]
\centering
\includegraphics[scale=0.38]{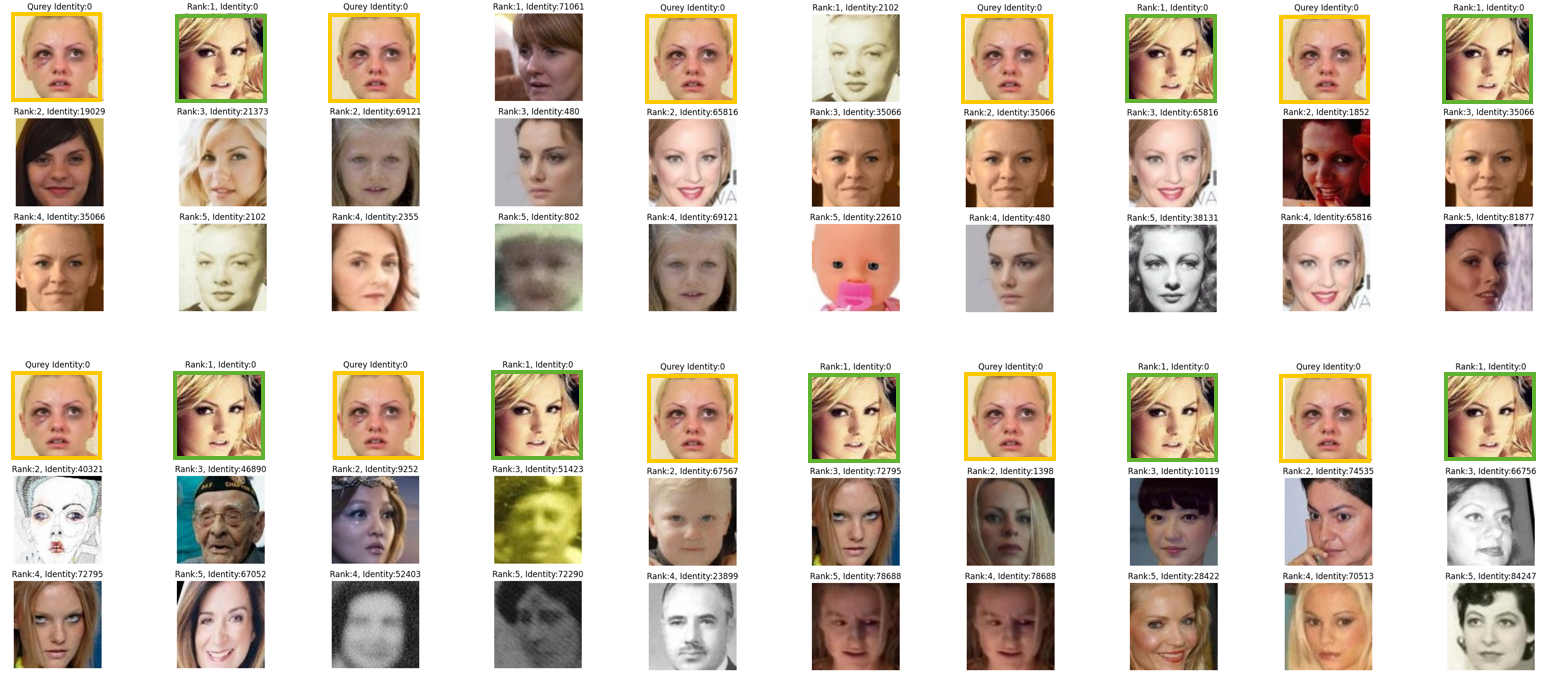}
\caption{Top 5 query results of  the hard-case in 1:N verification. The first three rows show the results based on original embeddings and last three rows present results based on protected embeddings. For left to right, the predictions are from ArcFace, DCTDP, DCTDP-FF, HFCF-LBP (concat) and HFCF-DLBP (concat), respectively. The query image is highlighted by yellow box while green box indicates the success prediction.}
\label{fig:singe_query}
\end{figure*}

To evaluate the identity-specific embedding mapping, a challenging query image is chosen for inference on various models. The results of the query are displayed in Figure \ref{fig:singe_query}.

The selected query image has a large dissimilarity compared with the image used for computing the enrolled embedding in the database. From Figure \ref{fig:singe_query}, DCTDP and DCTDP-FF both failed to recognize the query identity in the top 5 predictions on original embeddings. It shows that frequency information might not be enough for the model to extract discriminated information in the hard-case query image. While HFCF-LBP (concat) as well as HFCF-DLBP (concat) successfully recognize the query image even in the top 1 prediction. It is obvious that the hybrid frequency-color information captures useful visual information even it is in the form of a sparse representation. Here, we recommend HFCF-DLBP (concat) over HFCF-LBP (concat) since LBP features still contain much more visual structure compared with DLBP features. For the query results based on protected embeddings, all the methods successfully predicted the query image in the first place. Besides, it is interesting to observe that the other 4 predictions have quite high visual dissimilarity, especially for ArcFace and DCTDP. It can be the reason that embedding mapping enlarges the distance among different identities. Another good aspect of this feature is that it protects the privacy information of the query image by avoiding providing other high-visually similar identities. For example, in query results based on original embeddings, even if the correct identity is not shown in the top 5 predictions, we can still have a rough appearance perception by observing the first few predicted identities because the predictions are correlated with the visual information of the original RGB image. However, protected embeddings are computed based on identity-specific embedding mapping, which enlarges the distance among embeddings based on identity rather than visual similarity. Therefore, it is more difficult for people to have or ``guess'' the approximated appearance perception by observing query results when the system fails to recognize the query image. 

\subsubsection{Embedding Distributions}

\begin{figure}[hbt!]
     \centering
     \begin{subfigure}[b]{0.47\textwidth}
         \centering
         \includegraphics[width=\textwidth]{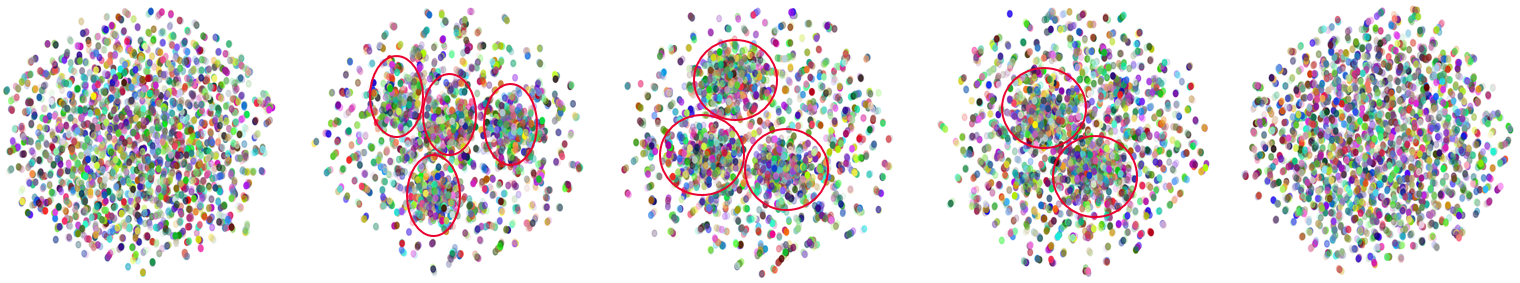}
         \caption{Original distribution.}
         \label{fig:a}
     \end{subfigure}
     \hfill
     \vspace*{-1mm}
     \begin{subfigure}[b]{0.47\textwidth}
         \centering
         \includegraphics[width=\textwidth]{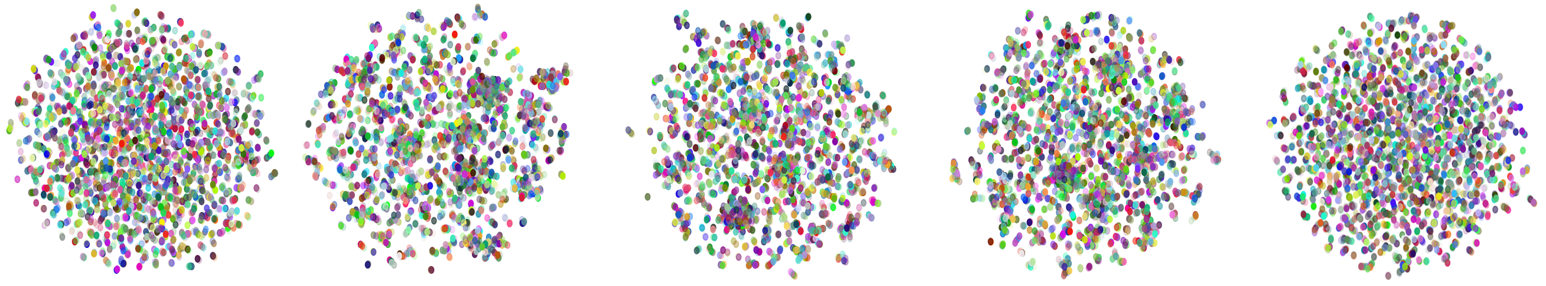}
         \caption{Protected distribution.}
         \label{fig:b}
     \end{subfigure}
        \caption{Distributions of original and protected face embeddings. From left to right, it shows results from ArcFace, DCTDP, DCTDP-FF, HFCF-LBP (concat), HFCF-DLBP (concat). The overlapped region is highlighted by red circle for better visualization.}
        \label{fig:Embedding distribution}
\end{figure}

In order to further verify the separation ability of protected embeddings, we select 3000 identities from the gallery dataset and compute the corresponding protected embeddings as well as the original embeddings. To visualize the distribution of high-dimensional embeddings, dimensionality reduction has to be applied. Uniform Manifold Approximation and Projection (UMAP) is chosen because it preserves both the local and global structures of the initial embeddings compared with t-distributed Stochastic Neighbor Embedding (t-SNE) \cite{van2008visualizing}. Basically, two groups that are separated in the embedded UMAP space are likewise far away in the original data. The visualization of embeddings is shown in Figure \ref{fig:Embedding distribution} by applying UMAP to reduce embeddings into two-dimensional representations.

For the original embedding distribution in Figure \ref{fig:a}, there are several heavily overlapped regions from DCTDP, DCTDP-FF and HFCF-LBP (concat) while such regions are not observed in ArcFace and HFCF-DLBP (concat). Besides, it is interesting to notice that the distribution from HFCF-DLBP (concat) is quite scattered in comparison with only the frequency-based method, which means the sparse color information introduced by DLBP can be learned by the feature extractor of the FR model. In the protected embedding distribution shown in Figure \ref{fig:b}, the previous large overlapped regions have decreased, even though there are still small congested areas. Therefore, the separation capability of the identity-specific embedding mapping scheme is manifest and obvious.

\section{\uppercase{Conclusion}}
In this paper, we have proposed a hybrid frequency-color fusion scheme named HFCF to convert RGB image data into a hybrid domain. There are two main features of HFCF: the dimensionality reduction in frequency information and the sparse visual representation in color information. HFCF can speed up the CNN training process and improve recognition accuracy with negligible color information. For face embedding protection, the identity-specific embedding mapping scheme with SMPC converts and securely calculates the embedding distance. Experimental results show that the proposed FR framework can yield excellent performance in both 1:1 and 1:N verification with good privacy preservation for input data as well as face embeddings. For the future work, we would like to investigate model robustness by performing black-box attacking through image reconstruction model. It is also interesting to see if proposed method can be generalized into other image contents in different tasks.

\bibliographystyle{apalike}
{\small
\bibliography{example}}



\end{document}